\documentclass[conference]{IEEEtran}
\IEEEoverridecommandlockouts
\usepackage{cite}
\usepackage{amsmath,amssymb,amsfonts}
\usepackage{algorithmic}
\usepackage{graphicx}
\usepackage{textcomp}
\usepackage[table,usenames,dvipsnames]{xcolor}
\usepackage{multirow}
\usepackage{booktabs}
\usepackage{hyperref}

\usepackage{makecell} % For multi-line cells
\def\BibTeX{{\rm B\kern-.05em{\sc i\kern-.025em b}\kern-.08em
    T\kern-.1667em\lower.7ex\hbox{E}\kern-.125emX}}

\usepackage{bm}

\usepackage{todonotes}
\setuptodonotes{inline}

\usepackage{xspace}

\newcommand{\seqtwor}{\texttt{\color{black}S2gR}\xspace}
\newcommand{\seqcap}{\texttt{\color{black}SCap}\xspace}
\newcommand{\seqtwo}{\texttt{\color{black}S2g}\xspace}
\newcommand{\seqonly}{\texttt{\color{black}SeqOnly}\xspace}

\begin{document}

% \title{Next Activity Prediction - A New Story from Resource-Perspective?\\
\title{Working My Way Back to You:\\ Resource-Centric Next-Activity Prediction\\
}

\author{\IEEEauthorblockN{Kelly Kurowski}
\IEEEauthorblockA{
\textit{Utrecht University}\\
Utrecht, The Netherlands \\
k.kurowski@uu.nl}
\and
\IEEEauthorblockN{Xixi Lu}
\IEEEauthorblockA{
\textit{Utrecht University}\\
Utrecht, The Netherlands \\
x.lu@uu.nl}
\and
\IEEEauthorblockN{Hajo A. Reijers}
\IEEEauthorblockA{
\textit{Utrecht University}\\
Utrecht, The Netherlands \\
h.a.reijers@uu.nl}
}

\maketitle
\thispagestyle{plain} %force page number to appear
\pagestyle{plain} %force page number to appear

\begin{abstract}
% Predictive Process Monitoring (PPM) enhances process mining by applying predictive models at runtime to forecast upcoming events during process execution. These predictions support early bottleneck detection, improved scheduling, proactive anomaly detection, and timely stakeholder communication. While existing work adopts a control-flow perspective, we investigate next-activity prediction from a resource-centric viewpoint, which offers additional benefits such as improved workload balancing and capacity forecasting. Although resource information has been shown to enhance tasks like resource allocation and process analysis, its role in next-activity prediction remains unexplored. In this study, we evaluate four prediction models and three encoding strategies across four real-life datasets. Compared to the baseline, our results show that LightGBM performs best with an encoding based on 2-gram activity transitions, while Random Forest and Transformer models benefit most from an encoding that combines 2-gram transitions and activity repetition features. This combined encoding also achieves the highest average accuracy. The findings underscore the potential of resource-centric next-activity prediction and suggest new opportunities and challenges for PPM research.

Predictive Process Monitoring (PPM) aims to train models that forecast upcoming events in process executions. These predictions support early bottleneck detection, improved scheduling, proactive interventions, and timely communication with stakeholders. While existing research adopts a control-flow perspective, we investigate next-activity prediction from a resource-centric viewpoint, which offers additional benefits such as improved work organization, workload balancing, and capacity forecasting. Although resource information has been shown to enhance tasks such as process performance analysis, its role in next-activity prediction remains unexplored. In this study, we evaluate four prediction models and three encoding strategies across four real-life datasets. Compared to the baseline, our results show that LightGBM and Transformer models perform best with an encoding based on 2-gram activity transitions, while Random Forest benefits most from an encoding that combines 2-gram transitions and activity repetition features. This combined encoding also achieves the highest average accuracy. This resource-centric approach could enable smarter resource allocation, strategic workforce planning, and personalized employee support by analyzing individual behavior rather than case-level progression. The findings underscore the potential of resource-centric next-activity prediction, opening up new venues for research on PPM.

\end{abstract}


\section{Introduction}
% echte wereld probleem (waar komt de vraag vandaan?), bestaande oplossingen? state of art, contributie (precies de voordelen benoemen), hoe goed is de contrubtie? wie kan dit gebruiken?

Predictive Process Monitoring (PPM) extends process mining by using predictive models that can be applied at runtime during the execution of a business process. This technique can be used to predict the next activity of a running case to detect bottlenecks, aim for better scheduling, enable proactive case handling through early anomaly detection, and timely stakeholder updates. 

At the time of writing this paper, next-activity prediction had always been applied from a case perspective within the field of process mining. However, predicting the next activity from a resource perspective arguably offers different benefits that could further enhance process mining applications. By anticipating the upcoming actions of individual resources, organizations can improve workload balancing, forecast capacity requirements, and proactively resolve potential conflicts or inefficiencies. This perspective could enable more informed resource allocation, support real-time process optimization, and contribute to better compliance and risk mitigation, particularly since research shows that resource behavior can impact process outcomes \cite{Kim2022ResourceExperience}.

In this paper, we investigate next-activity prediction from a resource-centric perspective by evaluating the impact of different encoding strategies, prediction models, and datasets on accuracy. We compare three different encoding strategies against a baseline to determine which one performs best. Additionally, using four different models, we investigate which encoding–model combinations are most effective. We also analyze how these combinations perform across varying prefix lengths for each dataset. Finally, we compare prediction performance across datasets. 

Our findings indicate that encodings incorporating activity transition patterns and run-length features—such as the frequency and average duration of repeated activities—can substantially enhance prediction accuracy and significantly outperform the baseline, particularly for early- to mid-length prefixes. In some cases, these approaches achieved up to a +0.18 improvement in accuracy compared to the baseline. These results demonstrate that resource-centric next-activity prediction is both feasible and promising, opening new directions for research in PPM.

\begin{figure*}[!t]
  \centering
  \includegraphics[width=0.70\textwidth]{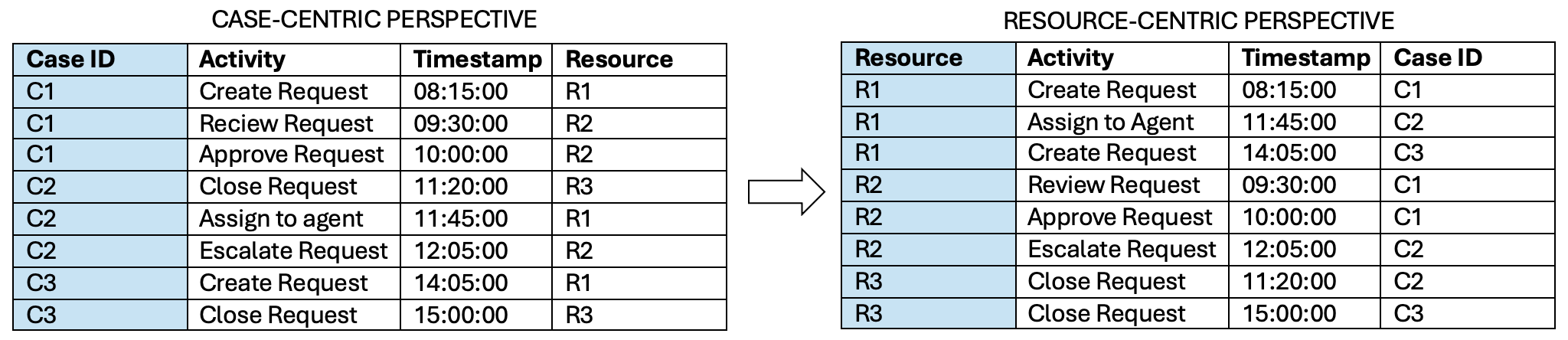}
  \caption{Case-centric perspective to resource-centric perspective event log}
  % \todo{Explain this figure in the text. Also good to add the targets.}
  \label{fig:your-label}
\end{figure*}

The remainder of this paper is structured as follows: Section \ref{sec:problem} outlines the importance and uniqueness of the problem; Section \ref{sec:related} discusses related work; Section \ref{sec:method} introduces the method and encodings; Section \ref{sec:results} presents the results; Section \ref{sec:discussion} summarizes key findings and presents new PPM opportunities; and Section \ref{sec:conclusion} concludes the paper.
% \cite{DBLP:journals/tist/VerenichDRMT19}

\section{Problem Description} \label{sec:problem}
One of the most common use cases in PPM is to predict the next activity of a process instance. Figure \ref{fig:your-label} illustrates how the event log is used for traditional next-activity prediction (left event log). By contrast, predicting the next activity from a \emph{resource-centric perspective} (right event log) has not yet been explored. We argue that this constitutes a new and valuable research direction, which is worthy of exploring for two reasons. 

\begin{table*}[h!]
  \centering
  \caption{Variant Ratio, Majority Class Prediction, and Example Leakage}
  \label{tab:majority_class}
  \resizebox{0.8\textwidth}{!}{
    \begin{tabular}{lcccccc}
      \toprule
      \textbf{Dataset} & \makecell{\textbf{Variant/Resource} \\ \textbf{Ratio}} & \makecell{\textbf{Variant/Case} \\ \textbf{Ratio}} & \makecell{\textbf{Resource-centric} \\ \textbf{Accuracy (avg.)}} & \makecell{\textbf{Case-centric} \\ \textbf{Accuracy (avg.)}} & \makecell{\textbf{Example Leakage} \\ \textbf{(Resource)}} & \makecell{\textbf{Example Leakage} \\ \textbf{(Case)}} \\
      \midrule
      BPIC 2013 Incident & 0.62 & 0.20 & 0.64 & 0.54 & 3\% & 83\% \\
      BPIC 2017           & 1.00 & 0.51 & 0.21 & 0.40 & 0.3\% & 80\% \\
      BPIC 2018           & 0.99 & 0.65 & 0.48 & 0.55 & 3\% & 12\% \\
      BPIC 2019           & 0.85 & 0.05 & 0.30 & 0.67 & 14\% & 85\% \\
      \bottomrule
    \end{tabular}
  }
\end{table*}

% From a \emph{motivational} perspective, this line of research can uncover patterns in how people work within organizations—specifically, whether they perform the same activities in long, repetitive sequences or in a more diverse and variable manner. Research suggests that individuals performing repetitive tasks are more prone to boredom and the accumulation of siloed knowledge \cite{vanHooff2016}. Such patterns can also reduce motivation and job satisfaction, potentially increasing turnover \cite{vanHooff2022}. Thus, by identifying repetitive sequences, organizations can better balance efficiency and resilience by encouraging task rotation or cross-training for resources. Moreover, repetitive sequences carried out by humans may be strong candidates for robotic process automation (RPA), thereby enhancing operational efficiency.

First, predicting the next activity of a resource opens up new possibilities for improving operational efficiency and workforce management. By anticipating what tasks a specific resource is likely to perform next, organizations could forecast workload more accurately and proactively balance task assignments \cite{renn2024employee}. This could allow better resource allocation, minimize bottlenecks, and reduce the risk of overloading key personnel. 
% Additionally, mature prediction techniques from a resource perspective could open new opportunities for PPM research. We present these opportunities in Section \ref{sec:discussion}.

Second, this presents a new and challenging problem. Predicting the next activity from a resource’s perspective presents unique challenges due to the high variability in task sequences and extremely long traces. We discuss this uniqueness by comparing three characteristics between case-centric and resource-centric views, as listed in Table~\ref{tab:majority_class}.  

In case-centric next-activity prediction, behavior is often dominated by a few majority variants. For example, the variant-to-case ratios of 0.20 in BPIC13 and 0.05 in BPIC19 suggest that, respectively, approximately only 20\% and 5\% of cases follow a unique sequence. In contrast, the resource-centric view shows significantly higher behavioral diversity. With the distinct variant-to-resource ratios of 1.00 in BPIC17 and 0.99 in BPIC19, this means (almost) every resource follows its own distinct execution sequence; see the first two columns in Table \ref{tab:majority_class}. This higher variability leads to much greater uncertainty.  
This means that the tasks performed by a given resource can vary widely, with many activities occurring much less frequently than others. 

To further illustrate the difficulty of resource-centric next-activity prediction, we evaluated the majority-class predictions across multiple prefix lengths for both resource-centric and case-centric next-activity prediction. We then calculated the average accuracy for each dataset, as presented in the fourth and fifth columns in Table \ref{tab:majority_class}. We observe higher majority-class prediction accuracy for case-centric next-activity prediction in the BPIC2017, BPIC2018, and BPIC2019 datasets.

Additionally, we examine the extent of example leakage, which refers to identical prefixes appearing in both training and test sets. Prior work has shown that example leakage is prevalent in case-centric next-activity prediction~\cite{abb2024generalization}. 
In our comparison, resource-centric prediction exhibited significantly lower leakage rates, as seen in the final two columns of Table~\ref{tab:majority_class}. 
% , we observed much lower example leakage of resource-centric next activity prediction in comparison with the case-centric approach. Example leakage occurs when a prefix with the same sequence of activities appears in both the training and test sets. 
This further underscores the increased complexity and generalization challenge posed by the resource-centric setting and its uniqueness.
% may indicate that the resource-centric problem is more challenging to solve than case-based activity prediction, where example leakage appeared to be more common \cite{abb2024generalization}. 
The average example leakage across all tested prefix lengths for the four datasets is reported in Table \ref{tab:majority_class}. We applied the method proposed by \cite{abb2024generalization} to calculate the example leakage for BPIC2018 and BPIC2019, as these datasets were not included in their original analysis.

These insights highlight the need to assess the effectiveness of existing next-activity prediction techniques on this new, resource-centric problem. Moreover, they motivate the development of new features and models tailored to capture the nuanced dynamics and individualized patterns in resource-centric behavior.

% \subsection{\tocheck{Formal stuff}}

\section{Related Work} \label{sec:related}

This section positions our work within existing PPM research. We begin with next-activity prediction, followed by an overview of process mining studies that incorporate resource-related features.

\subsection{Next-Activity Prediction}
To our knowledge, and based on recent systematic literature reviews on PPM, next-activity prediction has consistently been approached from a control-flow perspective~\cite{silva2024predictive, neu2022systematic, DBLP:conf/bpm/Francescomarino18}. A recent benchmark and review by Rama et al.~\cite{rama2023deeplearning} also shows that newer techniques, such as LSTMs and other deep learning models, continue to adopt a case-centric perspective. We describe three examples from their study.

\begin{table*}[t]
  \centering
  \caption{Dataset Summary}
  \label{tab:dataset_summary}
  \resizebox{\textwidth}{!}{
    % \begin{tabular}{lccccccc}
    \begin{tabular}{lrrrrrrr}
      \toprule
      \textbf{Event Log Name} & \textbf{Cases} & \textbf{Events} & \textbf{Activities} & \textbf{Unique Resources} & \makecell{\textbf{Avg. Activity Sequence} \\ \textbf{Length per Resource}} & \makecell{\textbf{Avg. Activity} \\ \textbf{Specialization per Resource}} & \makecell{\textbf{Avg. Activity} \\ \textbf{Repetition per Resource}} \\
      \midrule
      BPIC2013 Incidents & 7,554 & 65,533 & 13 & 1,440 & 40.51 & 0.34 & 43.27 \\
      BPIC2017           & 31,509 & 1,160,405 & 26 & 149   & 8,068.91 & 0.31 & 8,052.70 \\
      BPIC2018           & 43,809 & 2,514,266 & 170 & 165   & 15,237.97 & 0.39 & 15,226.57 \\
      BPIC2019           & 251,734 & 1,595,923 & 42 & 628   & 2,541.28 & 0.78 & 2,538.45 \\
      \bottomrule
    \end{tabular}
  }
\end{table*}

Tax et al. predict the next activity and timestamp using two functions \cite{tax2017predictive}. Each event is represented by a feature vector with a one-hot encoded activity type and time-based features. The prediction functions are modeled with LSTMs in different ways: separate models for each function, a joint multi-task LSTM model, or a hybrid approach with shared LSTM layers followed by specialized layers for each task.

In the work of Mehdiyev et al., predicting the next process event is treated as a classification problem \cite{mehdiyev2020novel}. The event log data is processed with a sliding window and encoded into n-grams, which are hashed and extended with data and resource features. Deep learning algorithms are used, starting with unsupervised pre-training to generate higher-level features, followed by supervised fine-tuning for multiclass classification. Evermann et al. present an early application of deep learning, specifically recurrent neural networks, for next process event prediction \cite{evermann2017deep}.

As discussed, to the best of our knowledge, no existing work has addressed next-resource prediction or examined next-activity prediction from a resource-centric perspective.

\subsection{Resource-Centric Process Mining}
Recently, more works have incorporated resource features or adopted more resource-centric approaches in process mining research \cite{klijn2024decomposing, kunkler2024online, kirchdorfer2024agentsimulator, DBLP:conf/bpm/TourPKS23}.
The work of Klijn et al. is resource-centric in the sense that it analyzes resource waiting time behavior to better understand process performance \cite{klijn2024decomposing}. The study shows that the features derived from a loan application process provide clearly interpretable performance insights compared to potentially misleading average waiting times, which do not account for resource behavior. 

Kunkler et al. showed that incorporating information about the estimated time a resource takes to perform a task can outperform traditional allocation strategies \cite{kunkler2024online}. This shows that incorporating information at the human or resource level allows for more nuanced and accurate process analysis.

Additionally, previous business process simulation (BPS) approaches followed a control-flow-first perspective by enriching process models with simulation parameters. In \cite{kirchdorfer2024agentsimulator}, the authors propose a resource-first approach that discovers a multi-agent system from an event log, capturing distinct resource behaviors and interactions. Their experiments show that this method achieves state-of-the-art accuracy with lower computation times and high interpretability across various process scenarios. Similarly, Tour et al. propose Agent Miner \cite{DBLP:conf/bpm/TourPKS23}, an algorithm for discovering models of agents and their interactions from event data. Their evaluation shows that the discovered models describe business processes more faithfully than those obtained using conventional process discovery algorithms.

These studies show that resource-feature can improve classical process mining tasks. We blend in with these streams by studying the next activity problem from a resource-centric perspective.

\section{Methodology} \label{sec:method}
To evaluate the effectiveness of various encoding strategies and prediction models for next-activity prediction from a resource-centric perspective, we conducted an experiment. We begin by presenting the objectives of the experiment, followed by a description of the datasets used. Next, we detail the encoding strategies, introduce the prediction models, and conclude with the evaluation measures.

\subsection{Objectives}
The objective of the experiment is to explore the differences in predictive performance (i.e., accuracy scores) between (1) different encoding strategies, (2) various prediction models, and (3) multiple datasets. Additionally, we evaluate these factors for varying prefix lengths and investigate whether incorporating additional features can improve prediction accuracy. This comparison aims to identify which combinations of encodings and models are most effective in capturing process behavior from a resource-centric perspective. 

\subsection{Datasets}
We examined all BPIC datasets to identify those with at least 100 unique resources. This threshold ensures that our test set included at least 20 distinct resources. Based on this criterion, we selected the following four publicly available real-life event logs: BPIC2013 Incidents, BPIC2017, BPIC2018, and BPIC2019.

Table \ref{tab:dataset_summary} summarizes the main properties of the selected logs, with a focus on resource characteristics, as these are particularly relevant to this research.
The average activity sequence length per resource is calculated by dividing the total number of activities performed per resource by the number of unique resources in the event log.

\emph{Activity specialization} reflects how narrowly a resource focuses on certain activities. It is based on Shannon entropy of the activity distribution per resource, normalized and inverted to yield a score between 0 (generalist) and 1 (specialist). The final value is the average specialization score across all resources.

The average activity repetition per resource captures how often resources repeat the same activities. For each resource, repetition is measured by counting how many times each activity occurs beyond its first appearance. This total is then divided by the number of unique activities in that sequence to get an average per resource, which is then averaged across all resources.

\subsection{Encoding Strategies}
We transformed each event log by sorting all activities by resource and timestamp, ensuring that each resource’s actions are represented in chronological order. For each resource, a sequence of activities was generated based on a specified prefix length, with the constraint that the resource must have at least as many activities as the prefix length. We evaluated multiple prefix lengths for each event log. However, the number of prefixes tested varied between logs. This variation occurred because we stopped increasing the prefix length once the number of resources dropped below 100, aiming to ensure at least 20 resources were included in the test set. We used a label encoder for each activity sequence. 
\subsubsection{Baseline Encoding- \seqonly} We use \seqonly as a baseline. For this encoding we predicted the next activity based solely on a sequence of activities, without providing any additional information to the model.
\subsubsection{\seqcap (capability)} The \seqcap encoding incorporated information about whether a resource has ever performed a specific activity within the entire event log. We used binary encoding for this purpose. Each activity from the event log was added as a separate column to the sequence data. A value of 1 indicated that the resource can perform the activity, while 0 signified that the resource cannot perform it. While we acknowledge that, from a classical machine learning perspective, this might be considered information leakage, we believe that in a business or organizational context it is always possible to infer what activities an employee can perform based on job descriptions, for instance.
\subsubsection{\seqtwo (2-gram)}
For our third encoding, which we refer to as \seqtwo, we investigated how 2-gram activity transitions would impact the prediction performance. For each possible transition, we calculated its frequency of occurrence in the observed prefix and incorporated this information into the model. We used a feature selection technique that selects the 20 most relevant features based on mutual information to reduce the number of columns. This method evaluates the relationship between each feature and the target variable, keeping only the top 20 features that provide the most useful information for prediction.
\subsubsection{\seqtwor (repetition)}
In the final encoding, \seqtwor, we introduced two new features focused on activity runs, tailored for the resource-centric perspective, added upon the previous encoding. First, for each resource, we calculated the number of consecutive runs of the same activity in the sequence. We then computed the average length of these runs by dividing the total length of all runs by the number of runs. A high value for this metric suggests that a resource stays in the same activity for longer periods, while a low value indicates more frequent switching between activities.

\subsection{Prediction Models}
We used four different prediction models: Random Forest, LightGBM, LSTM, and Transformer. We chose to include LightGBM instead of XGBoost because LightGBM is faster and newer \cite{bentejac2021comparative}.

For Random Forest and LightGBM, we applied GridSearchCV with cross-validation to tune key parameters such as the number of trees, the minimum number of samples required to split a node, the minimum number of samples required at a leaf node, and the maximum depth of the trees. For the LSTM and Transformer models, we use fixed configurations. These include preset values for learning rate, hidden size, number of layers, and the patience value for early stopping.

The LSTM model consists of two stacked LSTM layers, each with 50 units, followed by a dense output layer. The dense layer uses a softmax activation function to output predicted class probabilities for multi-class classification. The model is compiled with the Adam optimizer and categorical cross-entropy loss; early stopping is applied to prevent overfitting. 

The Transformer-based model represents each activity as a 128-dimensional embedding and employs a multi-layer Transformer encoder with 2 layers and 4 attention heads, each with a hidden size of 128. Positional encoding is added to preserve the order of activities in the sequence; the output is pooled and passed through a fully connected layer with output size matching the number of activity classes. The model is trained using Cross-Entropy Loss and the Adam optimizer (with a learning rate of 0.001) for 200 epochs, with early stopping applied if there is no improvement after 20 epochs. An overview of the hyperparameters tested per model can be found in Table \ref{tab:hyp_params}.

\begin{table}[h]
  \centering
  \caption{Hyperparameters for Random Forest, LightGBM, LSTM, and Transformer models}
  \label{tab:hyp_params}
  \scriptsize
  \begin{tabular}{lccccc}
    \toprule
    \textbf{Model}       & \textbf{Hyperparameter}          & \textbf{Possible Values}                              \\
    \midrule
    \multirow{5}{*}{Random Forest}  & n\_estimators                 & [50, 100, 200, 300]                                 \\
                          & max\_depth                    & [None, 10, 20, 30]                                  \\
                          & min\_samples\_split           & [2, 5, 10]                                          \\
                          & min\_samples\_leaf            & [1, 2, 4]                                           \\
                          & bootstrap                     & [True, False]                                       \\
    \midrule
    \multirow{5}{*}{LightGBM}     & n\_estimators                 & [50, 100, 200]                                      \\
                          & max\_depth                    & [-1, 10, 20]                                        \\
                          & learning\_rate                & [0.05, 0.1]                                         \\
                          & subsample                     & [0.8, 1.0]                                          \\
                          & colsample\_bytree             & [0.8, 1.0]                                          \\
    \midrule
    \multirow{3}{*}{LSTM}         & Units                         & 50                                                   \\
                          & Dropout                       & 0.2                                                 \\
                          & Optimizer                     & Adam                                                 \\
    \midrule
    \multirow{5}{*}{Transformer}  & d\_model                      & 128                                                  \\
                          & num\_heads                    & 4                                                    \\
                          & num\_layers                   & 2                                                    \\
                          & dropout                       & 0.1                                                  \\
                          & learning\_rate                & 0.001                                                \\
    \bottomrule
  \end{tabular}
  
\end{table}

To address class imbalance, any class that appears only once is either duplicated (if it is the only rare class), or merged into a generic placeholder class when multiple rare classes are present. The dataset is split using stratified train-test splitting (80/20) to preserve the original class distribution.

For transparency, all code used for this research, including dataset analyses, is available at \href{https://github.com/Kelly-Kurowski/PPM_ResourceCentric}{github.com/Kelly-Kurowski/PPM\_ResourceCentric}.

\setlength{\tabcolsep}{5pt}  % Reduce column spacing
{
\begin{table*}[t]
% \todo{Might be good to add another table, just report the actual accuracy. }
\caption{Average improved performance and standard deviation across different models, grouped by encoding, compared to the baseline.}
\label{tab:performance_diff}
\resizebox{\textwidth}{!}{
\centering
% \resizebox{\textwidth}{!}{%
\Huge
\begin{tabular}{lcccc|cccc|cccc}
\toprule
        & \multicolumn{4}{c|}{\Huge{\seqcap}} & \multicolumn{4}{c|}{\Huge{\seqtwo}} & \multicolumn{4}{c}{\Huge{\seqtwor}} \\ [4pt]
        & LightGBM & LSTM & Random Forest & Transformer & LightGBM & LSTM & Random Forest & Transformer & LightGBM & LSTM & Random Forest & Transformer \\
\midrule
BPIC13  & $0.000 \pm 0.010$ & $-0.001 \pm 0.006$ & $+0.006 \pm 0.019$ & $-0.001 \pm 0.064$ & $\text{\textbf{+0.076}} \pm 0.095$ & $+0.051 \pm 0.122$ & $+0.039 \pm 0.084$ & $+0.041 \pm 0.103$ & $+0.069 \pm 0.098$ & $\text{\textbf{+0.062}} \pm 0.121$ & $\text{\textbf{+0.072}} \pm 0.084$ & $\text{\textbf{+0.076}} \pm 0.114$ \\
BPIC17  & $-0.013 \pm 0.035$ & $-0.064 \pm 0.047$ & $-0.003 \pm 0.046$ & $-0.021 \pm 0.141$ & $+0.165 \pm 0.105$ & $+0.018 \pm 0.058$ & $+0.178 \pm 0.099$ & $\text{\textbf{+0.238}} \pm 0.178$ & $\text{\textbf{+0.182}} \pm 0.126$ & $\text{\textbf{+0.036}} \pm 0.059$ & $\text{\textbf{+0.187}} \pm 0.093$ & $-0.005 \pm 0.159$ \\
BPIC18  & $-0.008 \pm 0.041$ & $-0.047 \pm 0.032$ & $0.000 \pm 0.043$ & $+0.035 \pm 0.089$ & $\text{\textbf{+0.145}} \pm 0.172$ & $-0.052 \pm 0.049$ & $+0.061 \pm 0.077$ & $+0.054 \pm 0.123$ & $+0.037 \pm 0.127$ & $-0.054 \pm 0.046$ & $\text{\textbf{+0.164}} \pm 0.077$ & $\text{\textbf{+0.079}} \pm 0.090$ \\
BPIC19  & $-0.005 \pm 0.015$ & $-0.125 \pm 0.063$ & $+0.002 \pm 0.019$ & $+0.002 \pm 0.017$ & $\text{\textbf{+0.022}} \pm 0.077$ & $-0.084 \pm 0.043$ & $\text{\textbf{+0.032}} \pm 0.024$ & $+0.016 \pm 0.029$ & $\text{\textbf{+0.022}} \pm 0.061$ & $-0.088 \pm 0.046$ & $+0.023 \pm 0.032$ & $\text{\textbf{+0.032}} \pm 0.018$ \\
\bottomrule
\end{tabular}%
}
\end{table*}
}

\begin{table*}[t]
\caption{Average model accuracy and standard deviation across different encodings, grouped by model. Best-performing encoding-model combination per dataset (row-wise) is bolded.} 

\label{tab:accuracy_by_model}
\resizebox{\textwidth}{!}{
\centering
\Huge
\setlength{\tabcolsep}{12pt}
\begin{tabular}{l|ccc|ccc|ccc|ccc}
\toprule
        & \multicolumn{3}{c|}{\Huge{LightGBM}} & \multicolumn{3}{c|}{\Huge{LSTM}} & \multicolumn{3}{c|}{\Huge{Random Forest}} & \multicolumn{3}{c}{\Huge{Transformer}} \\ [4pt]
        & \seqcap & \seqtwo & \seqtwor & \seqcap & \seqtwo & \seqtwor & \seqcap & \seqtwo & \seqtwor & \seqcap & \seqtwo & \seqtwor \\
\midrule
BPIC13  & $0.65 \pm 0.06$ & $0.73 \pm 0.12$ & $0.72 \pm 0.13$
       & $0.68 \pm 0.03$ & $0.73 \pm 0.13$ & $0.74 \pm 0.13$
       & $0.67 \pm 0.04$ & $0.71 \pm 0.10$ & $\mathbf{0.74} \pm 0.09$
       & $0.59 \pm 0.10$ & $0.63 \pm 0.15$ & $0.66 \pm 0.10$ \\
BPIC17  & $0.45 \pm 0.10$ & $0.63 \pm 0.07$ & $0.64 \pm 0.09$
       & $0.28 \pm 0.05$ & $0.37 \pm 0.07$ & $0.38 \pm 0.08$
       & $0.46 \pm 0.08$ & $0.64 \pm 0.07$ & $\mathbf{0.65} \pm 0.05$
       & $0.36 \pm 0.08$ & $0.62 \pm 0.11$ & $0.37 \pm 0.09$ \\
BPIC18  & $0.49 \pm 0.14$ & $0.64 \pm 0.06$ & $0.53 \pm 0.08$
       & $0.43 \pm 0.05$ & $0.43 \pm 0.05$ & $0.42 \pm 0.05$
       & $0.49 \pm 0.09$ & $0.55 \pm 0.08$ & $\mathbf{0.65} \pm 0.05$
       & $0.53 \pm 0.13$ & $0.55 \pm 0.09$ & $0.58 \pm 0.12$ \\
BPIC19  & $0.88 \pm 0.08$ & $0.91 \pm 0.02$ & $0.91 \pm 0.02$
       & $0.61 \pm 0.08$ & $0.65 \pm 0.02$ & $0.64 \pm 0.02$
       & $0.89 \pm 0.03$ & $0.92 \pm 0.02$ & $0.91 \pm 0.03$
       & $0.89 \pm 0.03$ & $0.90 \pm 0.03$ & $\mathbf{0.92} \pm 0.02$ \\
\bottomrule
\end{tabular}%
}
\end{table*}

\subsection{Evaluation Measures}
We used the accuracy\_score function from the sklearn.metrics library to evaluate how well each model predicts the next activity of a resource. This metric calculates the proportion of correct predictions divided by the total number of predictions. The resulting accuracy scores are then compared across different models and encoding strategies for each event log.

% \todo{As far as I remembered, it was not a simple accuracy, it was a multi-class accuracy of sklearn? Maybe good to add the formula here.}

\section{Results} \label{sec:results}
\begin{figure*}[t]
    \centering
    \resizebox{\textwidth}{0.6\textheight}{%
        \includegraphics{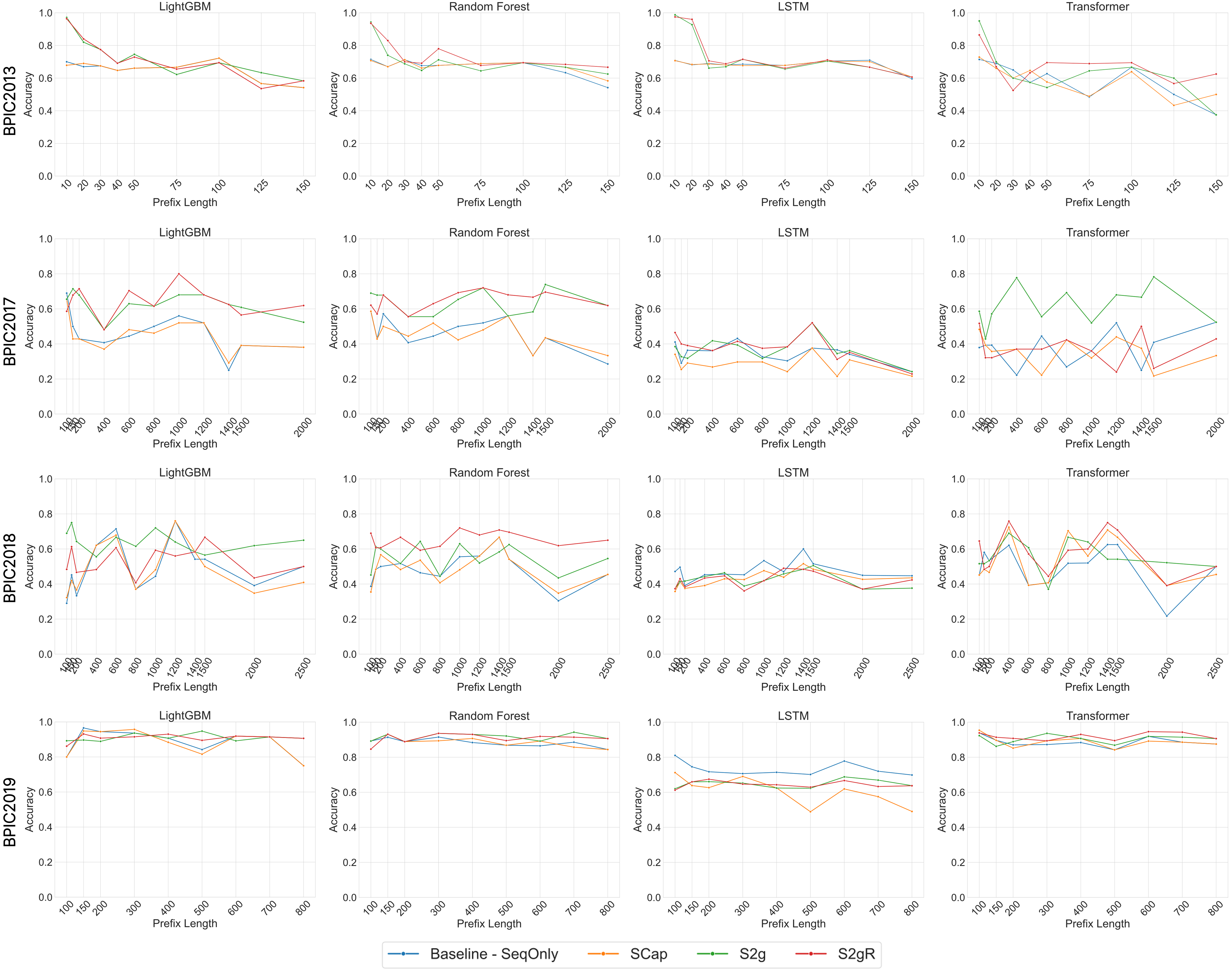}
    }
    \caption{Accuracy comparison of different models and encoding strategies across datasets and prefix lengths. A black-and-white version of this figure is available at:
\url{https://github.com/Kelly-Kurowski/PPM_ResourceCentric}, in the \textit{Final Graphs} directory.
}
    \label{fig:BPIC2013_Graph}
\end{figure*}

In this section, we present our results, organised by the objectives of the experiment. We begin by describing the differences in encoding strategies, followed by a similar analysis of the prediction models and, finally, the datasets.

\subsection{Performance of Encoding Strategies}

Figure~\ref{fig:BPIC2013_Graph} provides a detailed view of accuracy performance across different prefix lengths, per dataset and per model. Overall, we observe that both \seqtwo (green) and \seqtwor (red) achieve similar accuracy scores and often exceed the baseline (blue) and \seqcap (orange). 

%(1) our \seqtwo and \seqtwor improvement
We see notable performance improvements when using \seqtwo and \seqtwor at several prefix lengths. 
%
% (1.1) This happens in a few cases: BPIC13 early prefix
For instance, for BPIC2013 Incidents, we observe a substantial increase in accuracy at early prefix lengths (such as 10 and 20) for Random Forest, LightGBM, and LSTM models. At a prefix length of 10, the accuracy improves by approximately 0.25 over the baseline. 
This performance peak at prefix 10 coincides with a high example leakage rate (52.00\%), compared to an average of 6.36\% across all prefixes. Nevertheless, it is important to note that this increase in accuracy is not observed in the baseline or the \seqcap encoding, indicating that \seqtwo and \seqtwor are better at capturing the predictive patterns, even when potential leakage exists.
% By contrast, \seqtwo and \seqtwor may inadvertently capture and leverage leaked patterns more effectively, leading to an artificial performance boost at this specific prefix length.

% (1.2) They avoid the very low drops. 
In addition to these gains, the \seqtwo and \seqtwor encodings also help mitigate significant performance drops seen in the baseline. For example, in BPIC2017 at prefix length 1400 with LightGBM, the baseline achieves an accuracy score of only 0.25, whereas \seqtwo and \seqtwor reach 0.63. Similarly, at prefix length 2000 with Random Forest, accuracy increases from 0.29 (baseline) to 0.62 (\seqtwo and \seqtwor). This trend is not limited to a single dataset: in BPIC2018 at prefix length 100 with Random Forest, accuracy increases from 0.39 (baseline) to 0.69 (\seqtwor), and with the Transformer model at prefix length 2000, we observe a gain of approximately +0.30.
% the baseline achieves an accuracy score of 0.39, while \seqtwor reaches 0.69. Additionally, for the same dataset with the Transformer model at prefix length 2000, we observe a performance improvement of approximately +0.30. 
These results highlight that the \seqtwo and \seqtwor encodings can substantially compensate for the poorer performance of the baseline at certain prefix lengths.

In contrast, the \seqcap encoding, which captures resource capability, generally performs close to the baseline and occasionally underperforms. This suggests that capability-based representations alone may not be sufficient to support accurate next-activity prediction in complex, resource-centric settings.

% 2. Overall improvements 
Table~\ref{tab:performance_diff} summarizes the average performance difference of each encoding compared to the baseline, reported per model and dataset. Positive values indicate improved performance, while negative values indicate decreased performance. Compared to the baseline, \seqtwo achieves improved average accuracy in 14 out of 16 settings, while \seqtwor achieves better performance in 13 out of 16 cases. 

Bold values highlight the best-performing encoding for each dataset–model combination. For instance, a value of \textbf{+0.076} for BPIC13 with LightGBM under \seqtwo indicates that \seqtwo outperforms all other encodings (i.e., \seqcap, \seqtwor, and \seqonly) for that specific dataset and model. An exception is the LightGBM model on the BPIC2019 event log, where \seqtwo and \seqtwor perform equally well. 

Overall, the highest number of bold values is observed for the \seqtwor encoding, indicating that it generally outperforms the other methods, including the baseline, across most models and datasets. Specifically, \seqtwor outperforms \seqtwo in 9 settings, while \seqtwo outperforms \seqtwor in 4 settings. 
% Note that we count a win for \seqtwo for the LSTM model on the BPIC2019 dataset because, although it performs 0.084 worse than the baseline, it still outperforms \seqtwor, which has a performance of -0.088. 
% Compared to the baseline, \seqtwor achieves better performance in 13 out of 16 cases. 

That \seqtwor is the best-performing encoding becomes even more evident in Table \ref{tab:accuracy_by_model}, which reports the average accuracy per encoding, grouped by the model. For each data set, the bold value denotes the highest accuracy achieved. Across all data sets, \seqtwor consistently achieved the highest average accuracy, where three out of four data sets used Random Forest, and one used Transformer.

\subsection{Performance of Prediction Models}
% The Transformer model produced the highest peak accuracy in several datasets, notably in BPIC2017 with \seqtwo. However, its accuracy varied substantially with prefix length and encoding, especially in BPIC2017 and BPIC2018.

% The Random Forest model demonstrated consistently strong and stable performance across most datasets. It maintained a relatively high accuracy with minimal fluctuation, particularly for BPIC2018 and BPIC2019 when using \seqtwor compared to other encodings.

% LightGBM achieved moderate to high accuracy, with some variability depending on the encoding. In several cases (e.g., BPIC2018), it showed competitive performance compared to Random Forest and Transformer, especially for mid-range prefix lengths.

% The LSTM model generally produced lower accuracy than the other models, especially for longer prefix lengths. Although performance remained relatively stable across encodings, its overall accuracy was below that of the other models across most datasets.

% In summary, the Transformer model showed the highest accuracy peaks, Random Forest exhibited consistent performance across datasets, LightGBM offered a balance between the two, and LSTM lagged behind in overall performance.
% The results highlight notable differences in the behavior and effectiveness of the prediction models across datasets, encodings, and prefix lengths.

The Transformer model achieved the highest peak accuracies in several datasets at a certain prefix, particularly when combined with the \seqtwo encoding on more complex logs like BPIC2017. These peaks demonstrate the model’s ability to capture intricate sequential patterns. However, its performance was highly sensitive to both prefix length and encoding strategy—especially in BPIC2017 and BPIC2018—resulting in considerable variability. This suggests that while Transformers are powerful, they require careful tuning to maintain consistent performance across different conditions.

By contrast, the Random Forest model exhibited consistently strong and stable performance across most datasets and prefix lengths. Its accuracy remained relatively high with low fluctuation, especially when using the \seqtwor encoding on datasets like BPIC2018 and BPIC2019. It also achieved the highest average accuracy for three out of four data sets, see Table~\ref{tab:accuracy_by_model}. This robustness indicates that Random Forests are less affected by variability in the resource-centric sequences and may benefit from the aggregated statistical features provided by the encoding. 

The LightGBM model consistently delivered moderate to high accuracy, occupying a middle ground between the stability of Random Forest and the peak performance of Transformer. In some cases, such as BPIC2018, LightGBM achieved competitive results, particularly at mid-range prefix lengths. While its performance did vary with encoding, it generally maintained a good trade-off between model complexity and predictive robustness.

Finally, the LSTM model generally lagged behind the others in overall accuracy. Although its performance was relatively stable across encodings, it struggled to match the flexibility and power of the other models. This may be attributed to challenges in handling long sequences or to the limited complexity of the dataset features, which may not align well with what LSTM architectures typically require to excel, such as richer contextual dependencies seen in NLP tasks.

In summary, Transformer models offer the highest potential accuracy but are sensitive to variations, Random Forest provides the most consistent and robust performance, LightGBM balances accuracy and stability well, and LSTM underperforms in this specific context of resource-centric next-activity prediction.

\subsection{Performance across Datasets}
As shown in Figure \ref{fig:BPIC2013_Graph}, each horizontal lane depicts how four prediction models perform per dataset. Among these, the models achieve the highest overall scores for the BPIC2019 data set, with accuracy ranging approximately from 0.85 to 0.95, except for the LSTM model, which performs slightly lower, with scores between 0.70 and 0.80. This strong performance for BPIC2019 may be attributed to its high average activity-specialization per resource (0.78), as reported in Table~\ref{tab:dataset_summary}. This number indicates that individual resources in BPIC2019 tend to perform a narrower and more consistent set of activities, making their behavior more predictable, consequently, enhancing the model's ability to anticipate subsequent activities. 

% The baseline also shows relatively stable accuracy scores across different prefix lengths in BPIC2019. 
By contrast, the other data sets, i.e., BPIC2013 Incidents, BPIC2017, and BPIC2018, show a larger variability in predictive performance. For instance, in BPIC2018 with the Transformer model, accuracy ranges from as low as 0.20 to as high as 0.62. The reduced stability in BPIC2017 and BPIC2018 may be attributed to several factors, such as their low average activity-specialization per resource (0.31 and 0.39, respectively), long activity sequences, and high levels of activity repetition. Although BPIC2013 also has low specialization (0.34), its shorter cases and smaller activity space likely lead to more consistent and predictable performance.
These observations suggest that average activity-specialization and average sequence length per resource may serve as important indicators of how well resource-centric next-activity prediction models are likely to perform on a given dataset. 

\section{Discussion} \label{sec:discussion}
We summarize the key findings of our results, discuss the limitations of this research, and propose several opportunities for resource-centric PPM.

\subsection{Prediction Performance Summary}
Our experiments, which evaluate three encoding strategies across four predictive models and four event logs, reveal several important insights. Among the evaluated encodings, both \seqtwo and \seqtwor
% —based on 2-gram activity transitions like \seqtwo, but enriched with features capturing the number and average length of consecutive activity runs per resource—
outperformed the baseline, with \seqtwor emerging as the most effective overall and \seqtwo ranked second, as presented in the previous section. These results suggest that encoding 2-gram activity transitions, central to both \seqtwo and \seqtwor, is a promising method for next-activity prediction from a resource-centric perspective. 
% Specifically, \seqtwor achieved the highest improvement in accuracy in 9 out of 16 dataset-model combinations, compared to 5 for \seqtwo. 
% the highest accuracy in 13 out of 16 dataset–model combinations. 
More specifically, %\seqtwor outperformed both \seqcap and \seqtwo. 
\seqtwor achieves the best performance for all four data sets, and achieves the highest improvements against baseline in 9 out of 16 dataset–model settings, compared to 4 for \seqtwo (and one tie). As \seqtwor extends \seqtwo by incorporating features that capture the number and average length of consecutive runs of the same activity in the sequence per resource, this result suggests that these two newly proposed features are particularly useful for modeling individual resource behavior and anticipating their next activities.

In contrast, \seqcap, which only includes resource capability as a feature, performed similarly to the baseline and, in some cases, worse. This suggests that while capability-based features (despite that it is computed on the full data set) provide whether a resource \emph{can} execute a certain task or not, they may be insufficient for capturing the dynamics required for accurate prediction.

% When analyzing encoding–model combinations, \seqtwor performs best on average with the Random Forest model, delivering strong and stable accuracy across datasets and prefix lengths. In contrast, \seqtwo achieves its highest average accuracy when combined with LightGBM and Transformer models.

% Both \seqtwo and \seqtwor also mitigate the sharp accuracy drops observed with the baseline at longer prefix lengths. For example, in BPIC2017, the LightGBM model's accuracy at prefix length 1400 improves from 0.25 (baseline) to approximately 0.63 when using either encoding. Similar gains are observed for the Random Forest model at prefix length 2000, and across BPIC2018 with various models.

The analysis of prediction models shows distinct strengths and limitations across architectures. Transformer models achieve the highest peak accuracies, especially with \seqtwo on complex datasets, showing a lot of potential for this new resource-centric perspective. However, their performance is highly variable and sensitive to the settings, which may need more fine-tuning. Random Forest models offer the most consistent and robust accuracy across datasets, particularly when combined with \seqtwor. LightGBM performs comparably to Random Forest. LSTM models underperform overall, suggesting limited suitability for resource-centric next-activity prediction in their current form.

Dataset characteristics also significantly impact model effectiveness. BPIC2019 yields the highest overall accuracy, ranging from 0.85 to 0.95 for most models, which is likely due to its high average activity-specialization per resource (0.78), which suggests consistent and highly specialized resource behavior. In contrast, datasets with lower specialization rate, such as BPIC2017, BPIC2018, and BPIC2013, show greater variability and generally lower predictive performance. This is probably influenced by longer activity sequences and higher levels of repetition. 

Overall, our results show that with the appropriate encoding and models, prediction accuracy in resource-centric settings can be improved substantially, by up to +0.25 in cases such as BPIC2013 (prefix 10 and prefix 150) and BPIC2017, and even +0.30 in BPIC2018 prefix 2000, relative to the baseline. When revising the majority class prediction baseline (see Table~\ref{tab:majority_class}), the gains are even more striking; for example, in BPIC2019, accuracy increases of +0.62 (from 0.30 to 0.92). These findings confirm that, despite the diversity and complexity of resource behavior, considerable improvement can be achieved, suggesting unexplored potential to further improve performance through carefully designed representations and learning strategies tailored to resource-centric prediction tasks. 

\subsection{Limitations}
% \emph{Limitations:}
We acknowledge that not using hyperparameter tuning for the LSTM and Transformer models is a limitation of our current work. We did this primarily due to the long training times of these architectures. In future work, we aim to explore tuning strategies to potentially improve their performance.
% Additionally, we relied solely on accuracy as an evaluation metric; incorporating additional metrics such as precision, recall, and F1-score could offer a more comprehensive assessment of model effectiveness.

\subsection{Opportunities}
We see several opportunities for resource-centric next-activity prediction. By shifting the predictive focus from case progression to individual resource behavior, organizations can create new ways to support operational efficiency, workforce planning, and employee empowerment.
\subsubsection{Resource Allocation and Scheduling}
Resource-centric activity predictions could enable more intelligent and dynamic allocation of tasks. By knowing which activity and when a resource is likely to perform next, process managers can proactively assign tasks that align with expected behavior, availability, and skillsets. This could reduce idle time, minimizes task-switching overhead, and help balance workload distribution across the workforce.
\subsubsection{Forecasting Role Demand}
Predicting the future activities of individual resources could help identify patterns in task execution across roles. Aggregating these predictions over time allows organizations to forecast demand for specific roles or competencies. This supports strategic workforce planning by highlighting which roles may require scaling, retraining, or redistribution, especially in fast-changing or seasonal business environments.
\subsubsection{Personalized Recommendations and Prescriptions}
Accurate next-activity predictions can serve as the foundation for building personalized task or work schedule recommender systems for employees. Research suggests that individuals performing repetitive tasks are more prone to boredom and the accumulation of siloed knowledge \cite{vanHooff2016}. Such patterns can also reduce motivation and job satisfaction, potentially increasing turnover \cite{vanHooff2022}. Incorporating these heuristics in such a recommender system, employees could be supported by context-aware suggestions tailored to their work history and current workload. This can help reduce monotony or cognitive load, improve decision-making, and lead to more satisfying work experiences by aligning recommendations with individual preferences and strengths. 
% \subsubsection{Work Pattern Analysis}
% This line of research can uncover patterns in how people work within organizations. Research suggests that individuals performing repetitive tasks are more prone to boredom and the accumulation of siloed knowledge \cite{vanHooff2016}. Such patterns can also reduce motivation and job satisfaction, potentially increasing turnover \cite{vanHooff2022}. Thus, by identifying repetitive sequences, organizations can better balance efficiency and resilience by encouraging task rotation or cross-training for resources.

\section{Conclusion} \label{sec:conclusion}

In this study, we explored next-activity prediction from a resource-centric perspective. We have shown evidence that this problem does not exhibit significant example leakage, unlike the traditional NAP~\cite{abb2024generalization}, and is not dominated by majority class label prediction. 
Prior work has suggested that incorporating resource-specific features or adopting a resource-centered focus can enhance process mining tasks such as resource allocation and process simulation. We followed up on this and showed that resource-centric next-activity prediction is a novel and technically challenging problem with distinct benefits.

We evaluated three encoding strategies across four prediction models and four real-life datasets. Compared to a baseline using only activity sequences, encodings incorporating resource-aware patterns—such as activity transitions and run-length features—significantly improved prediction accuracy. Notably, the combination of Random Forest and \seqtwor yielded the best overall performance, while \seqtwo was most effective with LightGBM and Transformer models. By contrast, \seqcap, which incorporates resource capability information, performed similarly to the baseline and occasionally worse. We found that dataset characteristics, such as average activity specialization, can impact resource-centric predictive performance.

Our contributions are threefold: (1) a comprehensive empirical evaluation of resource-centric next-activity prediction using multiple models and encoding strategies on real-life datasets; (2) a demonstration that encoding activity transition patterns and run-length features substantially improves prediction accuracy; and (3) an introduction of a novel research perspective by shifting from case-centric to resource-centric prediction.

Our findings suggest promising directions for future work, including investigating the effect of more granular transition encoding and exploring applications in predicting resource or role demand. Overall, this research introduces the resource-centric next-activity prediction with valuable implications for enhancing resource management and process efficiency.

\bibliographystyle{IEEEtran}
\bibliography{ref}

\end{document}